\documentclass{article}

\PassOptionsToPackage{numbers, sort&compress}{natbib}
\usepackage[preprint]{neurips_2021}

\usepackage[utf8]{inputenc} 
\usepackage[T1]{fontenc}    
\usepackage{hyperref}       
\usepackage{url}            
\usepackage{booktabs}       
\usepackage{amsfonts}       
\usepackage{nicefrac}       
\usepackage{microtype}      
\usepackage{caption}
\usepackage{xcolor}         
\usepackage[pdftex]{graphicx}  
\usepackage{blindtext}      
\usepackage{algorithm}      
\usepackage{wrapfig}        

\usepackage{times}
\usepackage[utf8]{inputenc} 
\usepackage[T1]{fontenc}    
\usepackage{url}            
\usepackage{booktabs}       
\usepackage{nicefrac}       
\usepackage{microtype}      
\usepackage{graphics,color}

\usepackage{algorithm}
\usepackage{algpseudocode}
\usepackage{enumitem}

\usepackage{amsfonts}       
\usepackage{amsmath}       
\usepackage{amssymb}

\newcommand{\fig}[1]{Fig.~\ref{#1}}    
\newcommand{\tab}[1]{Table~\ref{#1}}

\renewcommand{\sec}[1]{Sec.~\ref{#1}} 
\newcommand{\supp}[1]{Suppl.~\ref{#1}}

\usepackage{xspace}
\makeatletter
\DeclareRobustCommand\onedot{\futurelet\@let@token\@onedot}
\def\@onedot{\ifx\@let@token.\else.\null\fi\xspace}
\def\eg{e.g\onedot}

\makeatother

\DeclareMathOperator*{\argmin}{arg\,min}

\usepackage{xcolor}
\definecolor{ourblue}{rgb}{0.368,0.507,0.71}
\definecolor{ourorange}{rgb}{0.881,0.611,0.142}
\definecolor{ourgreen}{rgb}{0.56,0.692,0.195}
\definecolor{ourred}{rgb}{0.923,0.386,0.209}
\definecolor{ourviolet}{rgb}{0.528,0.471,0.701}
\definecolor{ourbrown}{rgb}{0.772,0.432,0.102}
\definecolor{ourlightblue}{rgb}{0.364,0.619,0.782}
\definecolor{ourdarkgreen}{rgb}{0.572,0.586,0.}

\definecolor{ourcyan2}{rgb}{0.125,0.722,0.804}
\definecolor{ourred2}{rgb}{0.863,0.184,0.047}
\definecolor{ouryellow2}{cmyk}{0,0.16,1.0,0.07}
\definecolor{ourviolet2}{cmyk}{0.55,0.56,0,0.47}
\definecolor{ourorange2}{cmyk}{0,0.46,0.89,0.11}

\title{Planning from Pixels in Environments with Combinatorially Hard Search Spaces}

\author{
  Marco Bagatella \\
  Max Planck Institute for Intelligent Systems \\
  Tübingen, Germany \\
  \texttt{mbagatella@tue.mpg.de} \\
  \And
  Mirek Ol\v{s}\'ak\\
  Computer Science Department\\
  University Innsbruck, Austria\\
  \texttt{mirek@olsak.net} \\
  \And
  Michal Rol\'inek\\
  Max Planck Institute for Intelligent Systems \\
  Tübingen, Germany \\
  \texttt{michal.rolinek@tue.mpg.de} \\
  \And
  Georg Martius\\
  Max Planck Institute for Intelligent Systems \\
  Tübingen, Germany \\
  \texttt{georg.martius@tue.mpg.de} \\
}

\begin{document}

\maketitle

\begin{abstract}
  The ability to form complex plans based on raw visual input is a litmus test for current capabilities of artificial intelligence, as it requires a seamless combination of visual processing and abstract algorithmic execution, two traditionally separate areas of computer science.
  A recent surge of interest in this field brought advances that yield good performance in tasks ranging from arcade games to continuous control; these methods however do not come without significant issues, such as limited generalization capabilities and difficulties when dealing with combinatorially hard planning instances.
  Our contribution is two-fold: (i) we present a method that learns to represent its environment as a latent graph and leverages state reidentification to reduce the complexity of finding a good policy from exponential to linear (ii) we introduce a set of lightweight environments with an underlying discrete combinatorial structure in which planning is challenging even for humans. 
  Moreover, we show that our methods achieves strong empirical generalization to variations in the environment, even across highly disadvantaged regimes, such as ``one-shot'' planning, or in an offline RL paradigm which only provides low-quality trajectories.
\end{abstract}

\begin{center}
    \includegraphics[width=0.95\linewidth]{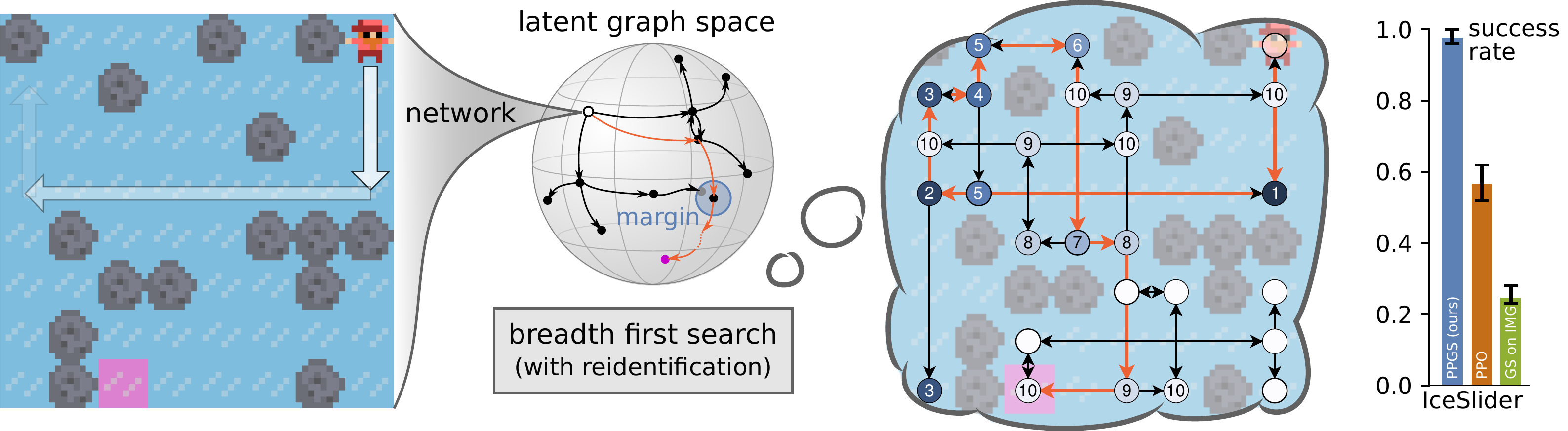}
    \captionof{figure}{Planning from Pixels with Graph Search. 
    Our method leverages learned latent dynamics to efficiently build and search a graph representation of the environment. Resulting policies show unrivaled performance across a distribution of hard combinatorial tasks.}
    \label{fig:overview}
\end{center}

\section{Introduction}

Decision problems with an underlying combinatorial structure pose a significant challenge for a learning agent, as they require both the ability to infer the true low-dimensional state of the environment and the application of abstract reasoning to master it.
A traditional approach for common logic games, given that a simulator or a model of the game are available, consists in applying a graph search algorithm to the state diagram, effectively simulating several trajectories to find the optimal one.
As long as the state space of the game grows at a polynomial rate with respect to the planning horizon, the solver is able to efficiently find the optimal solution to the problem.
Of course, when this is not the case, heuristics can be introduced at the expense of optimality of solutions.

Learned world models \cite{ha2018recurrent, hafner2018learning} can learn to map complex observations to a lower-dimensional latent space and retrieve an approximate simulator of an environment.
However, while the continuous structure of the latent space is suitable for training reinforcement learning agents \cite{hafner2019dream, chua2018pets} or applying heuristic search algorithms \cite{schrittwieser2019mastering}, it also prevents a straightforward application of simpler graph search techniques that rely on identifying and marking visited states.

Our work follows naturally from the following insight: a simple graph search might be sufficient for solving visually complex environments, as long as a world model is trained to realize a suitable structure in the latent space.
Moreover, the complexity of the search can be reduced from exponential to linear by reidentifying visited latent states.

The method we propose is located at the intersection between classical planning, representation learning and model-based reinforcement learning. 
It relies on a novel low-dimensional world model trained through a combination of opposing losses without reconstructing observations.
We show how learned latent representations allow a dynamics model to be trained to high accuracy, and how the dynamics model can then be used to reconstruct a \textit{latent graph} representing environment states as vertices and transitions as edges.
The resulting latent space structure enables powerful graph search algorithms to be deployed for planning with minimal modifications, solving challenging combinatorial environments from pixels. We name our method \textbf{\textsc{PPGS}} as it \textbf{P}lans from \textbf{P}ixels through \textbf{G}raph \textbf{S}earch.

We design \textsc{PPGS} to be capable of generalizing to unseen variations of the environment, or equivalently across a distribution of \textit{levels} \cite{cobbe2020procgen}. This is in contrast with traditional benchmarks \cite{bellemare2012ale}, which require the agent to be trained and tested on the same fixed environment.

We can describe the main contributions of this paper as follows:
first, we introduce a suite of environments that highlights a weakness of modern reinforcement learning approaches,
second, we introduce a simple but principled world model architecture that can accurately learn the latent dynamics of a complex system from high dimensional observations;
third, we show how a planning module can simultaneously estimate the latent graph for previously unseen environments and deploy a breadth first search in the latent space to retrieve a competitive policy;
fourth, we show how combining our insights leads to unrivaled performance and generalization on a challenging class of environments.

\section{Method}
\label{sec:method}
For the purpose of this paper, each environment can be modeled as a family of fully-observable deterministic goal-conditioned Markov Decision Processes with discrete actions, that is the 6-tuples $\{(S, A, T, G, R, \gamma)_i\}_{1...n}$ where $S_i$ is the state set, $A_i$ is the action set, $T_i$ is a transition function $T_i: S_i \times A_i \rightarrow S_i$, $G_i$ is the goal set and $R_i$ is a reward function $R_i: S_i \times G_i \rightarrow \mathbb{R}$ and $\gamma_i$ is the discount factor.
We remark that each environment can also be modeled as a BlockMDP \cite{du2019provably} in which the context space $\mathcal{X}$ corresponds to the state set $S_i$ we introduced.

In particular, we deal with families of procedurally generated environments. We refer to each of the $n$ elements of a family as a \textit{level} and omit the index $i$ when dealing with a generic level.
We assume that state spaces and action spaces share the same dimensionality across all levels, that is $|S_i|=|S_j|$ and $|A_i|=|A_j|$ for all $0 \leq i, j \leq n$.

In our work the reward simplifies to an indicator function for goal achievement $R(s, g) = \mathbf{1}_{s=g}$ with $G \subseteq S$. Given a goal distribution $p(g)$, the objective is that of finding a goal-conditioned policy $\pi_g$ that maximizes the return

\begin{align}
\mathcal{J}_{\pi} &= 
\displaystyle \mathop{\mathbb{E}}_{g \sim p(g)} \Bigg[ \mathop{\mathbb{E}}_{\tau \sim p(\tau | \pi_g)} \sum_t \gamma^t R(s_t,g) \Bigg]
\label{eq:return}
\end{align}

where $\tau \sim p(\tau | \pi_g)$ is a trajectory $(s_t, a_t)_{t=1}^T$ sampled from the policy.

Our environments of interest should challenge both perceptive and reasoning capabilities of an agent. In principle, they should be solvable through extensive search in hard combinatorial spaces.
In order to master them, an agent should therefore be able to (i) identify pairs of bisimilar states \cite{zhang2020learning}, (ii) keep track of and reidentify states it has visited in the past and (iii) produce highly accurate predictions for non-trivial time horizons.
These factors contribute to making such environments very challenging for existing methods.
Our method is designed in light of these necessities; it has two integral parts, the world model and the planner, which we now introduce.

\subsection{World Model}
\label{sec:world_model}
\begin{figure}
    \begin{center}
    \includegraphics[width=0.8\linewidth]{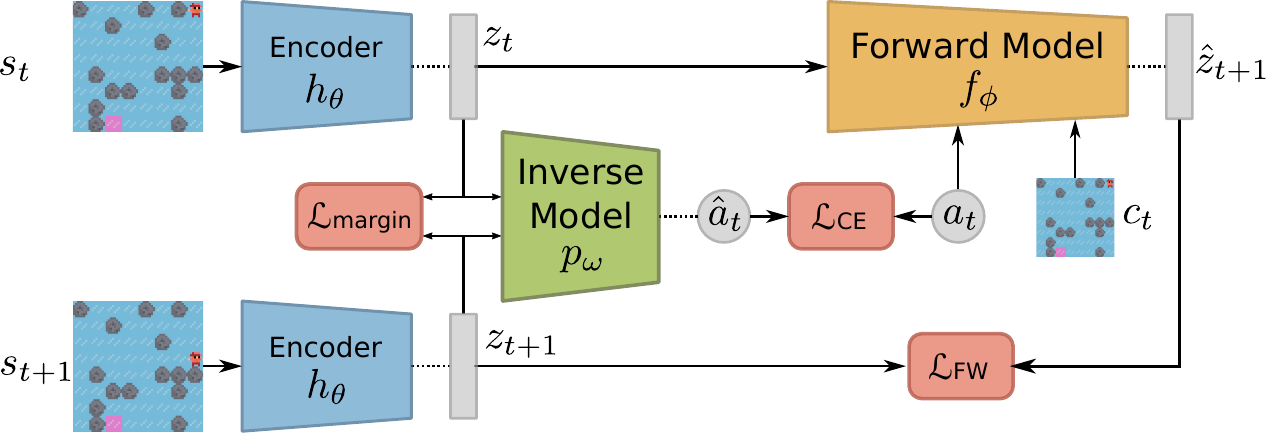}    
    \end{center}
    \caption{Architecture of the world model. A convolutional encoder extracts latent state representations from observations, while a forward model and an inverse model reconstruct latent dynamics by predicting state transitions and actions that cause them. The notation is introduced in \sec{sec:world_model}}
    \label{fig:model}
\end{figure}

The world model relies solely on three jointly trained function approximators: an encoder, a forward model and an inverse model. Their overall orchestration is depicted in \fig{fig:model} and described in the following.

\subsubsection{Encoder}

Mapping highly redundant observations from an environment to a low-dimensional state space $Z$ has several benefits \cite{ha2018recurrent, hafner2018learning}. Ideally, the projection should extract the compressed ``true state'' of the environment and ignore irrelevant visual cues, discarding all information that is useless for planning.
For this purpose, our method relies on an \textit{encoder} $h_\theta$, that is a neural function approximator mapping each observed state $s \in S$ and a low-dimensional representation $z \in Z$ (\textit{embedding}).
While there are many suitable choices for the structure of the latent space $Z$, we choose to map observations to points on an $d$-dimensional hypersphere taking inspiration from \citet{liu2017sphereface}.

\subsubsection{Forward Model}

In order to plan ahead in the environment, it is crucial for an agent to estimate the transition function $T$. In fact, if a mapping to a low-dimensional latent space $Z$ is available, learning directly the projected transition function $T_Z: Z \times A \to Z$ can be largely beneficial \cite{ha2018recurrent, hafner2018learning}.
The deterministic latent transition function $T_Z$ can be learned by a neural function approximator $f_\phi$ so that if $T(s_t, a_t) = s_{t+1}$, then $f_\phi(h_\theta(s_t), a_t) := f_\phi(z_t, a_t) = h_\theta(s_{t+1})$.
We refer to this component as \textit{forward model}. Intuitively, it can be trained to retrieve the representation of the state of the MDP at time $t+1$ given the representation of the state and the action taken at the previous time step $t$.

Due to the Markov property of the environment, an initial state embedding $z_t$ and the action sequence $(a_t, \dots, a_{t+k})$ are sufficient to to predict the latent state at time $t+k$, as long as $z_t$ successfully captures all relevant information from the observed state $s_t$.
The amount of information to be embedded in $z_t$ and to be retained in autoregressive predictions is, however, in most cases, prohibitive.
Take for example the case of a simple maze: $z_t$ would have to encode not only the position of the agent, but, as the predictive horizon increases, most of the structure of the maze.

\paragraph{Invariant Structure Recovery}
To allow the encoder to only focus on local information, we adopt an hybrid forward model which can recover the invariant structures in the environment from previous observations. The function that the forward model seeks to approximate can then include an additional input: $f_\phi(z_t, a_t, s_c) = z_{t+1}$, where $s_c \in S$ is a generic observation from the same environment and level.
Through this context input the forward model can retrieve information that is constant across time steps (\eg the location of walls in static mazes).
In practice, we can use randomly sampled observation from the same level during training and use the latest observation during evaluation.
This choice allows for more accurate and structure-aware predictions, as we show in the ablations in \supp{app:ablation}.

Given a trajectory $(s_t, a_t)_{t=1}^{T}$, the forward model can be trained to minimize some distance measure between state embeddings $(z_{t+1})_{1\ldots T-1} = (h_\theta(s_{t+1}))_{1\ldots T-1}$ and one-step predictions $(f_\phi(h_\theta(s_{t}), a_{t}, s_{c}))_{1\ldots T-1}$.
In practice, we choose to minimize a Monte Carlo estimate of the expected Euclidean distance over a finite time horizon, a set of trajectories and a set of levels. When training on a distribution of levels $p(l)$, we extract $K$ trajectories of length $H$ from each level with a uniform random policy $\pi$ and we minimize
\begin{align}
\mathcal{L}_{\text{FW}} &= 
\displaystyle \mathop{\mathbb{E}}_{l \sim p(l)} \bigg[
\frac{1}{H-1} \sum_{h=1}^{H-1}
\displaystyle \mathop{\mathbb{E}}_{a_h \sim \pi}
\Big[
\| f_\phi(z^l_{h}, a_h, s^l_c) - z^l_{h+1} \|_2^2
\Big]
\bigg]
\label{eq:loss-fw}
\end{align}
where the superscript indicates the level from which the embeddings are extracted.

\subsubsection{Inverse Model and Collapse Prevention}

Unfortunately, the loss landscape of Equation \ref{eq:loss-fw} presents a trivial minimum in case the encoder collapses all embeddings to a single point in the latent space.
As embeddings of any pair of states could not be distinguished in this case, this is not a desirable solution.
We remark that this is a known problem in metric learning and image retrieval \cite{bellet2013survey}, for which solutions ranging from siamese networks \cite{bromley1993signature} to using a triplet loss \cite{hoffer2015deep} have been proposed.

The context of latent world models offers a natural solution that isn't available in the general embedding problem, which consists in additionally training a probabilistic \textit{inverse model} $p_\omega(a_t \mid z_t, z_{t+1})$ such that if $T_Z(z_t, a_t) = z_{t+1}$, then $p_\omega(a_t \mid z_t, z_{t+1}) > p_\omega(a_k \mid z_t, z_{t+1}) \forall a_k \neq a_t \in A$.
The inverse model, parameterized by $\omega$, can be trained to predict the action $a_t$ that causes the latent transition between two embeddings $z_t, z_{t+1}$ by minimizing multi-class cross entropy.
\begin{align}
\mathcal{L}_{\text{CE}} =
\displaystyle \mathop{\mathbb{E}}_{l \sim p(l)} \bigg[
\frac{1}{H-1} \sum_{h=1}^{H-1}
\displaystyle \mathop{\mathbb{E}}_{a_h \sim \pi}
\Big[
- \log{p_\omega(a_h \mid z^l_{h}, z^l_{h+1})}
\Big]
\bigg].
\label{eq:loss-ce}
\end{align}
Intuitively, $\mathcal{L}_{\text{CE}}$ increases as embeddings collapse, since it becomes harder for the inverse model to recover the actions responsible for latent transitions. For this reason, it mitigates unwanted local minima. Moreover, it is empirically observed to enforce a regular structure in the latent space that eases the training procedure, as argued in \sec{app:ablation} of the Appendix.
We note that this loss plays a similar role to the reconstruction loss in \citet{hafner2018learning}. However, $\mathcal{L}_{CE}$ does not force the encoder network to embed information that helps with reconstructing irrelevant parts of the observation, unlike training methods relying on image reconstruction \cite{chiappa2017recurrent, ha2018recurrent, hafner2018learning, hafner2019dream, hafner2020mastering}.

While $\mathcal{L}_{\text{CE}}$ is sufficient for preventing collapse of the latent space, a discrete structure needs to be recovered in order to deploy graph search in the latent space. In particular, it is still necessary to define a criterion to reidentify nodes during the search procedure, or to establish whether two embeddings (directly encoded from observations or imagined) represent the same true low-dimensional state.

A straightforward way to enforce this is by introducing a margin $\varepsilon$, representing a desirable minimum distance between embeddings of non-bisimilar states \cite{zhang2020learning}. A third and final loss term can then be introduced to encourage margins in the latent space:
\begin{align}
\mathcal{L}_{\text{margin}} =
\displaystyle \mathop{\mathbb{E}}_{l \sim p(l)} \bigg[
\frac{1}{H-1} \sum_{h=1}^{H-1}
\max\Big(0, 1 - \frac{\|z^l_{h+1} - z^l_{h}\|_2^2}{\varepsilon^2}\Big)
\bigg].
\label{eq:loss-margin}
\end{align}
We then propose to reidentify two embeddings as representing the same true state if their Euclidean distance is less than $\frac{\varepsilon}{2}$.

Adopting a latent margin effectively constrains the number of margin-separated states that can be represented on an hyperspherical latent space.
However, this quantity is lower-bounded by the kissing number~\cite{kissing_number}, that is the number of non-overlapping unit-spheres that can be tightly packed around one $d$ dimensional sphere.
The kissing number grows exponentially with the dimensionality $d$. 
Thus, the capacity of our $d$-dimensional unit sphere latent space ($d=16$ in our case with margin $\varepsilon=0.1$) is not overly restricted. 

The world model can be trained jointly and end-to-end by simply minimizing a combination of the three loss functions:

\begin{align}
\mathcal{L} = \alpha \mathcal{L}_{\text{FW}} + \beta \mathcal{L}_{\text{CE}} + \mathcal{L}_{\text{margin}}.
\label{eq:loss-total}
\end{align}

To summarize, the three components are respectively encouraging accurate dynamics predictions, regularizing latent representations and enforcing a discrete structure for state reidentification.

\subsection{Planning Regimes}
\label{sec:planner}
\begin{figure}
    \includegraphics[width=\linewidth]{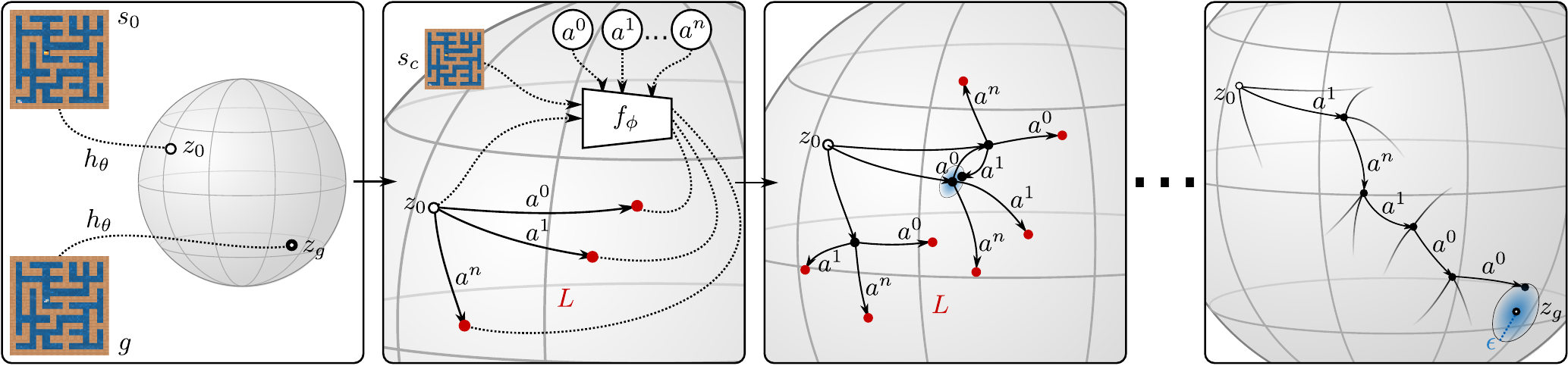}
    \caption{Overview of latent-space planning. One-shot planning is possible by (i) embedding the current observation and goal to the latent space and (ii) iteratively growing a latent graph until a vertex is reidentified with the goal.}
    \label{fig:tree}
\end{figure}

A deterministic environment can be represented as a directed graph $G$ whose vertices $V$ represent states $s \in S$ and whose edges $E$ encode state transitions. An edge from a vertex representing a state $s \in S$ to a vertex representing a state $s' \in S$ is present if and only if $T(s, a) = s'$ for some action $a \in A$, where $T$ is the state transition function of the environment. This edge can then be labelled by action $a$. Our planning module relies on reconstructing the \emph{latent graph}, which is a projection of graph $G$ to the latent state $Z$.

\setlength{\intextsep}{.2em}
\begin{wrapfigure}{r}{0.40\linewidth}
\centering
\includegraphics[width=\linewidth]{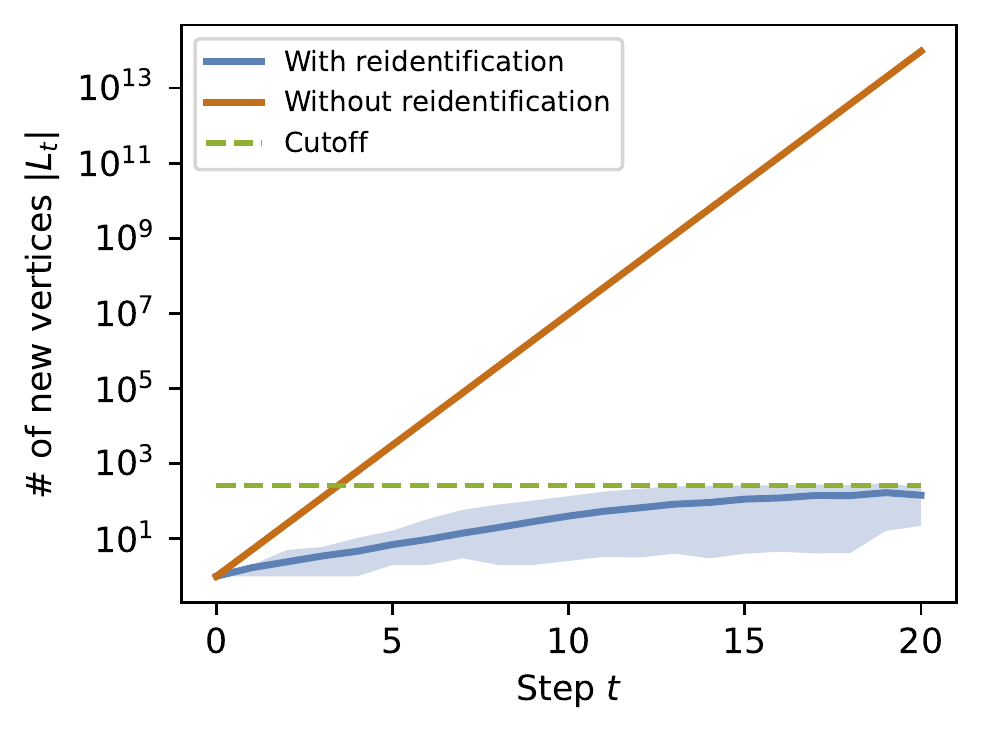}
\caption{Number of leaf vertices when planning in ProcgenMaze, averaged over 100 levels, with 90\% confidence intervals.}
\label{fig:fringe}
\end{wrapfigure}
\setlength{\intextsep}{1.0em}

In this section we describe how a latent graph can be build from the predictions of the world model and efficiently searched to recover a plan, as illustrated in \fig{fig:tree}. 
This method can be used as a one-shot planner, which only needs access to a visual goal and the initial observation from a level. When iterated and augmented with online error correction, this procedure results in a powerful approach, which we refer to as \emph{full planner}, or simply as \textsc{PPGS}.

\paragraph{One-shot Planner}
Breadth First Search (BFS) is a graph search algorithm that relies on a LIFO queue and on marking visited states to find an optimal path $O(V+E)$ steps.
Its simplicity makes it an ideal candidate for solving combinatorial games by exploring their latent graph.
If the number of reachable states in the environment grows polynomially, the size of the graph to search will increase at a modest rate and the method can be applied efficiently.

We propose to execute a BFS-like algorithm on the latent graph, which is recovered by autoregressively simulating all transitions from visited states. As depicted in \fig{fig:tree}, at each step, the new set of leaves $L$ is retrieved by feeding the leaves from the previous iteration through the forward model $f_\phi$.
The efficiency of the search process can be improved as shown in \fig{fig:fringe}, by exploiting the margin $\varepsilon$ enforced by equation $\ref{eq:loss-margin}$ to reidentify states and identify loops in the latent graph.
We now provide a simplified description of the planning method in Algorithm \ref{alg:simplified}, while details can be found in \supp{app:planner}.
\begin{algorithm}[htbp]
\caption{Simplified one-shot \textsc{PPGS}}
\textbf{Input:} Initial observed state $s_1$, visual goal $g$, model parameters $\theta, \phi$
\begin{algorithmic}[1]
\State $z_1, z_g = h_\theta(s_1), h_\theta(g)$ \Comment{project to latent space $Z$}
\State $L, V= \{z_1\}$ \Comment{sets of leaves and visited vertices}
\For{$T_{MAX}$ steps}
    \State $L = \{f_\phi(z, a, s_1) : \exists z \in L, a \in A\}$ \Comment{grow graph}
    \If{$z^* \in L$ can be reidentified with $z_g$}
        \State \textbf{return} action sequence from $z_1$ to $z^*$
    \EndIf
    \State $L = L \setminus V$ \Comment{reidentify and discard visited vertices (details in \supp{app:planner})}
    \State $V = V \cup L$ \Comment{update visited vertices}
\EndFor
\end{algorithmic}
\label{alg:simplified}
\end{algorithm}

\paragraph{Full Planner}
The one-shot variant of \textsc{PPGS} largely relies on highly accurate autoregressive predictions, which a learned model cannot usually guarantee.
We mitigate this issue by adopting a model predictive control-like approach \cite{garcia1989model}. \textsc{PPGS} recovers an initial guess on the best policy $(a_i)_{1, ..., n}$ simply by applying one-shot \textsc{PPGS} as described in the previous paragraph and in Algorithm \ref{alg:open-loop}.
It then applies the policy step by step and projects new observations to the latent space.
When new observations do not match with the latent trajectory, the policy is recomputed by applying one-shot \textsc{PPGS} from the latest observation.
This happens when the autoregressive prediction of the current embedding (conditioned on the action sequence since the last planning iteration) can not be reidentified with the embedding of the current observation.
Moreover, the algorithm stores all observed latent transitions in a lookup table and, when replanning, it only trusts the forward model on previously unseen observation/action pairs.
A detailed description can be found in \supp{app:planner}.

\section{Environments}
\label{sec:envs}

In order to benchmark both perception and abstract reasoning, we empirically show the feasibility of our method on three challenging procedurally generated environments. These include the Maze environment from the procgen suite \cite{cobbe2020procgen}, as well as DigitJump and IceSlider, two combinatorially hard environments which stress the reasoning capabilities of a learning agent, or even of an human player.
In the context of our work, the term “combinatorial hardness” is used loosely.
We refer to an environment as "combinatorially hard" if only very few of the exponentially many trajectories actually lead to the goal, while deviating from them often results in failure (e.g. DigitJump or IceSlider).
Hence, some “intelligent” search algorithm is required.
In this way, the process of retrieving a successful policy resembles that of a graph-traversing algorithm.
The last two environments are made available in a public repository \cite{environments}, where they can also be tested interactively.
More details on their implementation are included in \supp{app:environments}.

\begin{figure}[!htb]
\minipage{0.27\textwidth}\centering\small
    ProcgenMaze\\
  \includegraphics[width=\linewidth]{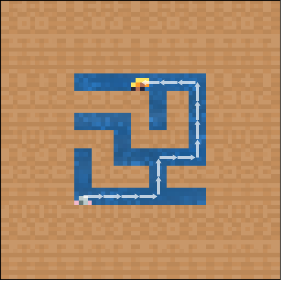}
\endminipage\hfill
\minipage{0.27\textwidth}\centering\small
DigitJump\\
  \includegraphics[width=\linewidth]{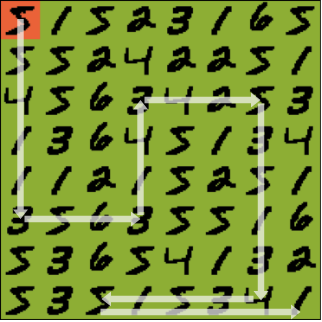}
\endminipage\hfill
\minipage{0.27\textwidth}\centering\small
    \ IceSlider\phantom{p}\\
  \includegraphics[width=\linewidth]{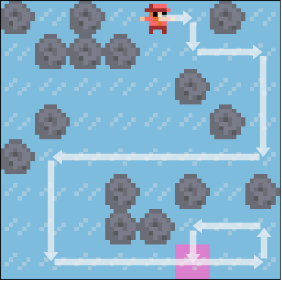}
\endminipage
\caption{Environments. Initial observations and one-shot \textsc{PPGS}'s solution (arrows) of a random level of each of the three environments. ProcgenMaze is from~\cite{cobbe2020procgen}. DigitJump and IceSlider are proposed by us and can be accessed at \cite{environments}.}
\end{figure}

\paragraph{ProcGenMaze}
The ProcgenMaze environment consists of a family of procedurally generated 2D mazes.
The agent starts in the bottom left corner of the grid and needs to reach a position marked by a piece of cheese.
For each level, an unique shortest solution exists, and its length is usually distributed roughly between $1$ and $40$ steps.
This environment presents significant intra-level variability, with different sizes, textures, and maze structures. While retrieving the optimal solution in this environment is already a non-trivial task, its dynamics are uniform and actions only cause local changes in the observations.
Moreover, ProcgenMaze is a forgiving environment in which errors can always be recovered from.
In the real world, many operations are irreversible, for instance, cutting/breaking objects, gluing parts, mixing liquids, etc.
Environments containing remote controls, for example, show non-local effects.
We use these insights to choose the additional environments.

\paragraph{IceSlider}
IceSlider is in principle similar to ProcgenMaze, since it also consists of procedurally generated mazes.
However, each action propels the agent in a direction until an obstacle (a rock or the borders of the environments) is met.
We generate solvable but unforgiving levels that feature irreversible transitions, that, once taken, prevent the agent from ever reaching the goal.

\paragraph{DigitJump}
DigitJump features a distribution of randomly generated levels which consist of a 2D 8x8 grid of handwritten digits from 1 to 6. The agent needs to go from the top left corner to the bottom right corner. The 4 directional actions are available, but each of them causes the agent to move in that directions by the number of steps expressed by the digit on the starting cell. Therefore, a single action can easily transport the player across the board. This makes navigating the environment very challenging, despite the reduced cardinality of the state space. Moreover, the game presents many cells in which the agent can get irreversibly stuck.

\section{Related Work}

\paragraph{World Models and Reinforcement Learning}
The idea of learning to model an environment has been widely explored in recent years.
Work by \citet{oh2015action} and \citet{chiappa2017recurrent} has argued that modern machine learning architectures are capable of learning to model the dynamics of a generic environment reasonably well for non-trivial time horizons.
The seminal work by \citet{ha2018recurrent} built upon this by learning a world model in a low-dimensional latent space instead of conditioning predictions on observations. They achieved this by training a VAE on reconstructing observations and a recurrent network for sampling latent trajectories conditioned on an action sequence. Moreover, they showed how sample efficiency could be addressed by recovering a simple controller acting directly on latent representations through an evolutionary approach.

This initial idea was iteratively improved along two main directions. On one hand, some subsequent works focused on learning objectives and suggested to jointly train encoding and dynamics components.
\citet{hafner2018learning} introduced a multi-step variational inference objective to encourage latent representations to be predictive of the future and propagate information through both deterministic and stochastic paths.
On the other hand, authors proposed to learn to act in the latent space by using zero-order methods \cite{hafner2018learning} such as CEM \cite{rubinstein1997optimization} or policy gradient techniques \cite{hafner2019dream, hafner2020mastering}. These improvements gradually led to strong model-based RL agents capable of achieving very competitive performance in continuous control tasks \cite{hafner2019dream} and on the Atari Learning Environment \cite{bellemare2012ale, chen2020simple, hafner2020mastering}.

Relying on image reconstruction can however lead to vulnerability to visual noise: to overcome this limitation \citet{okada2020dreaming} and \citet{zhang2020learning} forgo the decoder network, while the latter proposes to rely on the notion of bisimilarity to learn meaningful representations.
Similarly, \citet{gelada2019deepmdp} only learn to predict rewards and action-conditional state distributions, but only study this task as an additional loss to model-free reinforcement learning methods.
Another relevant approach is that of \cite{zhang2020world}, who propose to learn a discrete graph representation of the environment, but their final goal is that of recovering a series of subgoals for model-free RL.

A strong example of how world models can be coupled with classical planners is given by MuZero \cite{schrittwieser2019mastering}. MuZero trains a recurrent world model to guide a Monte Carlo tree search by encouraging hidden states to be predictive of future states and a sparse reward signal. While we adopt a similar framework, we focus on recovering a discrete structure in the latent space in order to reidentify states and lower the complexity of the search procedure. Moreover, we do not rely on reward signals, but only focus on learning the dynamics of the environment.

\paragraph{Neuro-algorithmic Planning}
In recent years, several other authors have explored the intersection between representation learning and classical algorithms. This is the case, for instance,
of \citet{kuo2018deep, kumar2019lego, ichter2018robot} who rely on sequence models or VAEs to propose trajectories for sampling-based planners.
Within planning research, \citet{yonetani2020path} introduce a differentiable version of the A* search algorithm that can learn suitable representations from images with supervision.
The most relevant line of work to us is perhaps the one that attempts to learn representations that are suitable as an input for classical solvers.
Within this area, \citet{asai2018classical, asai2020learning} show how symbolic representations can be extracted from complex tasks in an end-to-end fashion and directly fed into off-the-shelf solvers.
More recently, \citet{vlastelica2021neuroalgorithmic} frames MDPs as shortest-path problems and trains a convolutional neural network to retrieve the weights of a fixed graph structure.
The extracted graph representation can be solved with a combinatorial solver and trained end-to-end by leveraging the blackbox differentiation method \cite{Pogancic2020Differentiation}.

\paragraph{Visual Goals}
A further direction of relevant research is that of planning to achieve multiple goals~\cite{nair2018visual}. While the most common approaches involve learning a goal-conditioned policy with experience relabeling \cite{andrychowicz2017hidsight}, the recently proposed GLAMOR~\cite{paster2020planning} relies on learning inverse dynamics and retrieves policies through a recurrent network. By doing so, it can achieve visual goals without explicitly modeling a reward function, an approach that is sensibly closer to ours and can serve as a relevant comparison.
Another method that sharing a similar setting to ours is LEAP \cite{nasiriany2019planning}, which also attempts to fuse reinforcement learning and planning; however, its approach is fundamentally different and designed for dense rewards and continuous control.
Similarly, SPTM \cite{savinov2018semi} pursues a similar direction, but requires exploratory traversals in the current environment, which would be particularly hard to obtain due to procedural generation.

\section{Experiments}
\label{sec:experiments}

The purpose of the experimental section is to empirically verify the following claims: (i) \textsc{PPGS} is able to solve challenging environments with an underlying combinatorial structure and (ii) \textsc{PPGS} is able to generalize to unseen variations of the environments, even when trained on few levels. We aim to demonstrate that forming complex plans in these simple-looking environments is beyond the reach of the best suited state-of-the-art methods. Our approach, on the other hand, achieves non-trivial performance. With this in mind, we did not insist on perfect fairness of all comparisons, as the different methods have different type of access to the data and the environment. However, the largest disadvantage is arguably given to our own method.

While visual goals could be drawn from a distribution $p(g)$, we evaluate a single goal for each test level matching the environment solution (or the only state that would give a positive reward in a sparse reinforcement working framework).
This represents a very challenging task with respect to common visual goal achievement benchmarks \cite{paster2020planning}, while also allowing comparisons with reward-based approaches such as \textsc{PPO} \cite{shulman2017ppo}.
We mainly evaluate the success rate, which is computed as the proportion of solved levels in a set of 100 unseen levels.
A level is considered to be solved when the agent achieves the visual goal (or receives a non-zero reward) within 256 steps.

\paragraph {Choice of Baselines}
Our method learns to achieve visual goals by planning with a world model learned on a distribution of levels. To the best of our knowledge, no other method in the literature shares these exact settings.
For this reason, we select three diverse and strong baselines and we make our best efforts for a fair comparison within our computational limits.

\textsc{PPO} \cite{shulman2017ppo} is a strong and scalable policy optimization method that has been applied in procedurally generated environments \cite{cobbe2020procgen}.
While \textsc{PPGS} requires a visual goal to be given, \textsc{\textsc{PPO}} relies on a (sparse) reward signal specializing on a unique goal per level.
DreamerV2 \cite{hafner2020mastering} is a model-based RL approach that also relies on a reward signal, while \textsc{GLAMOR} \cite{paster2020planning} is more aligned with \textsc{PPGS} as it is also designed to reach visual goals in absence of a reward.

While we restrict \textsc{PPGS} to only access an offline dataset of low-quality random trajectories, all baselines are allowed to collect data on policy for a much larger number of environment steps.
More considerations on these baselines and on the fairness of our comparison can be found in \supp{app:fairness}.
Furthermore, we also consider a non-learning naive search algorithm  (\textsc{GS on Images}) thoroughly described in \ref{app:classical}.

A comprehensive ablation study of \textsc{PPGS} can be found in Section \ref{app:ablation} of the Appendix.

\setlength{\intextsep}{.2em}
\begin{wrapfigure}[16]{r}{.45\textwidth}
  \centering
    \includegraphics[width=\linewidth]{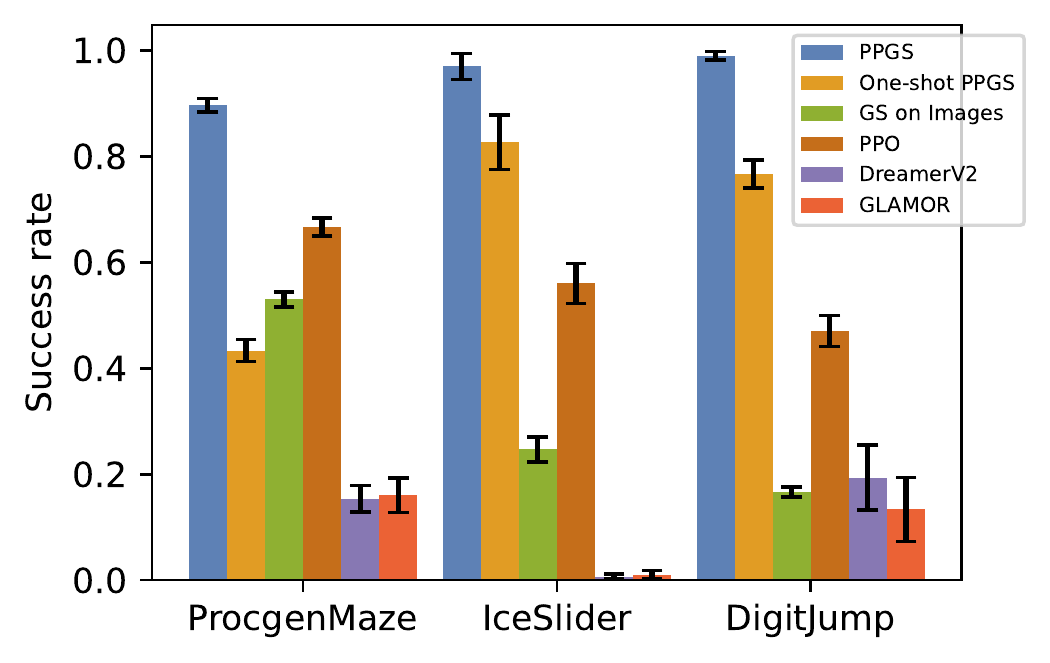}
    \caption{Success rates across the three environments. One-shot planning is competitive with the full method on shorter time horizons.}
    \label{fig:success}
\end{wrapfigure}
\setlength{\intextsep}{1.0em}

\subsection{Comparison of Success Rates}
Our first claim is supported by Figure \ref{fig:success}.
\textsc{PPGS} outperform its baselines across the three environments.
The gap with baselines is smaller in ProcgenMaze, a forgiving environment for which accurate plans are not necessary. On the other hand, ProcgenMaze involves long-horizon planning, which can be seen as a limitation to one-shot \textsc{PPGS}.
As the combinatorial nature of the environment becomes more important, the gap with all baselines increases drastically.

\textsc{PPO} performs fairly well with simple dynamics and long-term planning, but struggles more when combinatorial reasoning is necessary. \textsc{GLAMOR} and DreamerV2 struggle across the three environments, as they likely fail to generalize across a distribution of levels.
The fact that \textsc{GS on Images} manages to rival other baselines is a testament to the harshness of the environments.

\subsection{Analysis of Generalization}
\begin{figure}[b]
  \centering
  \begin{minipage}{0.45\textwidth}
  \centering
    \includegraphics[width=\textwidth]{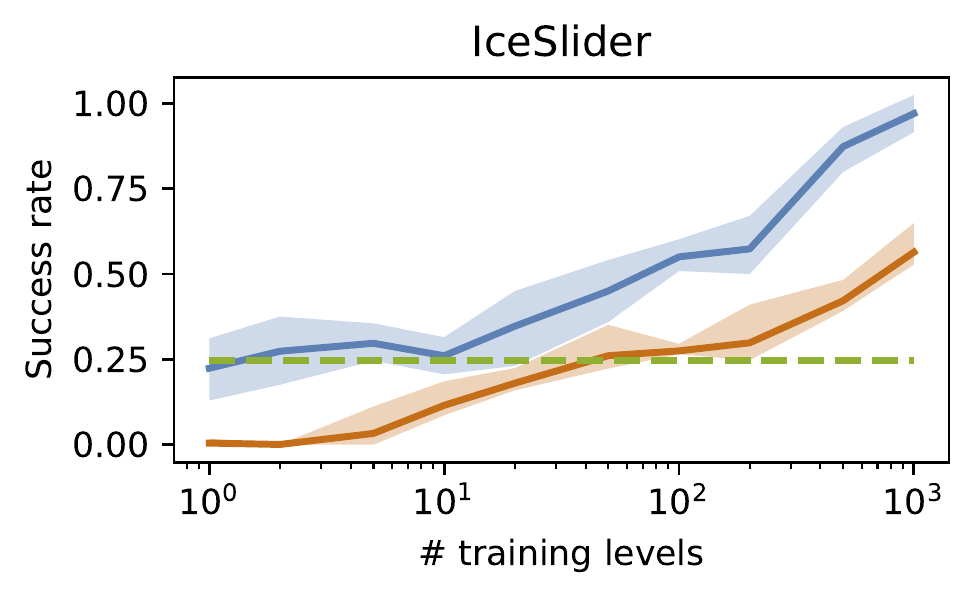}
  \end{minipage}
  \hfill
  \begin{minipage}{0.45\textwidth}
  \centering
    \includegraphics[width=\textwidth]{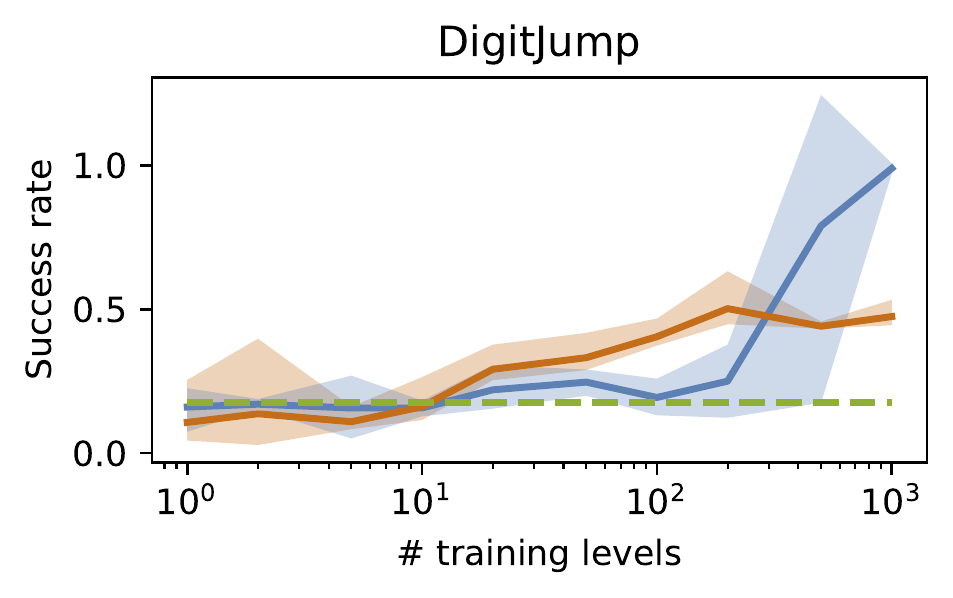}
  \end{minipage}\\[-.5em]
  \begin{minipage}{0.45\textwidth}
    \includegraphics[width=\textwidth]{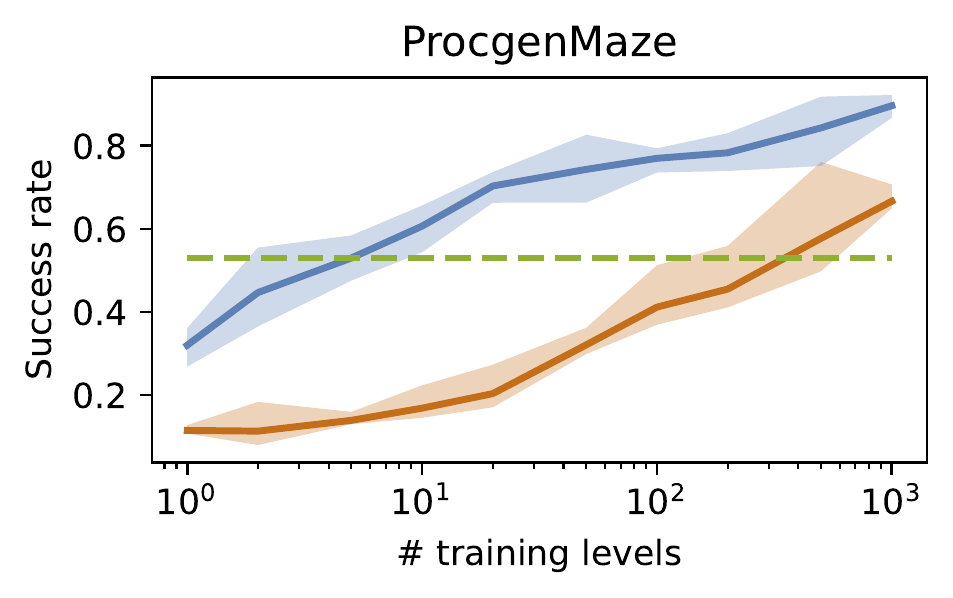}
  \end{minipage}
  \hfill
  \begin{minipage}{0.45\textwidth}
  \centering
      \includegraphics[width=0.4\textwidth]{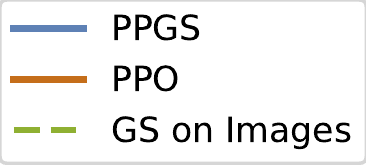}
  \end{minipage}
  \caption{Solution rates of \textsc{PPGS} and \textsc{PPO} as a function of the cardinality of the set of training levels.}
\label{fig:generalization}
\end{figure}

The inductive biases represented by the planning algorithm and our training procedure ensure good generalization from a minimal number of training levels.
In \fig{fig:generalization}, we compare solution rates between \textsc{PPGS} and \textsc{PPO} as the number of levels available for training increases. The same metric for larger training level sets is additionally available in \tab{tab:numerical}.
Our method generally outperforms its baselines across all environments.
In ProcgenMaze, \textsc{PPGS} achieves better success rates than \textsc{PPO} after only seeing two orders of magnitude less level, \eg 10 levels instead of 1000. Note that \textsc{PPGS} uses only 400k samples from a random policy whereas \textsc{PPO} uses 50M on-policy samples. 
Due to the harshness of the remaining environments, \textsc{PPO} struggles to find a good policy and its solution rate on unseen levels improves slowly as the number of training levels increases. In IceSlider, \textsc{PPGS} is well above \textsc{PPO} for any size of the training set and a outperforms \textsc{GS on Images} when only having access to $2$ training levels.
While having a comparable performance to \textsc{PPO} on small training sets in DigitJump, our method severely outperforms it once approximately $200$ levels are available. On the other hand, \text{PPO}'s ability to generalize plateaus.
These results show that \textsc{PPGS} quickly learns to extract meaningful representations that generalize well to unseen scenarios.

\section{Discussion}
\label{sec:discussion}

\paragraph{Limitations}
The main limitations of our method regard the assumptions that characterize the class of environments we focus on, namely a slowly expanding state space and discrete actions.
In general, due to the complexity of the search algorithms, scaling to very large action sets becomes challenging.
Moreover, a single expansion of the search tree requires a forward pass of the dynamics network, which takes a non-negligible amount of time.
Finally, the world model is a fundamental component and the accuracy of the forward model is vital to the planner.
Training an accurate forward model can be hard when dealing with exceedingly complex observations: very large grid sizes in the environments are a significant obstacle.
On the other hand, improvements in the world model would directly benefit the whole pipeline.

\paragraph{Conclusion}
Hard search from pixels is largely unexplored and unsolved, yet fundamental for future AI.
In this paper we presented how powerful graph planners can be combined with learned perception modules to solve challenging environment with a hidden combinatorial nature.
In particular, our training procedure and planning algorithm achieve this by (i) leveraging state reidentification to reduce planning complexity and (ii) overcoming the limitation posed by information-dense observations through an hybrid forward model.
We validated our proposed method, \textsc{PPGS}, across three challenging environments in which we found state-of-the-art methods to struggle. We believe that our results represent a sensible argument in support of the integration of learning-based approaches and classical solvers.

\begin{ack}
We acknowledge the support from the German Federal Ministry of Education and Research (BMBF) through the Tübingen AI Center (FKZ: 01IS18039B).
Georg Martius is a member of the Machine Learning Cluster of Excellence, funded by the Deutsche Forschungsgemeinschaft (DFG, German Research Foundation) under Germany’s Excellence Strategy – EXC number 2064/1 – Project number 390727645.

\end{ack}

\bibliographystyle{abbrvnat}
\bibliography{main}

\newpage
\appendix
\begin{center}
    \Large\bf  Supplementary Material for:\\
Planning from Pixels in Environments\\ with Combinatorially Hard Search Spaces
\end{center}
\section{Ablation Study}
\label{app:ablation}

\textsc{PPGS} relies on crucial architectural choices that we now set out to motivate. We do so by performing an ablation study and questioning each of the choices individually to show its contribution to the final performance.

\paragraph{World Model}
We evaluate the impact of different choices on the world model by retraining it from scratch and reporting the success rate of the full planner in \tab{tab:ablation}.
We also compute two latent metrics, which are commonly used to benchmark latent predictions \cite{kipf2019contrastive}, see below.

\begin{table*}[bh]
\small
\centering
\resizebox{\textwidth}{!}{%
\begin{tabular}{@{}lccccc@{}}
\toprule
& \multicolumn{5}{c}{\textbf{ProcgenMaze / DigitJump}}\\
\cmidrule{2-6} & 
Success \% & H@1 & H@10 & MMR@1 & MMR@10\\
\midrule
Our method & 0.91/1.00 & 1.00/1.00 & 0.92/0.99 & 1.00/1.00  & 0.94/1.00  
\\ \midrule
\quad without inverse model & 0.38/0.23 & 1.00/1.00 & 0.77/0.23 & 1.00/1.00  & 0.81/0.33
\\
\quad with fully latent forward model & 0.80/0.34 & 0.98/1.00 & 0.73/0.53 & 0.98/1.00  & 0.81/0.63  
 
\\ \midrule
\quad without lookup table & 0.39/0.92 & - & - & - & -
\\
\quad one-shot, with reidentification & 0.43/0.76 & - & - & - & -
\\
\quad one-shot, without reidentification & 0.27/0.31 & - & - & - & -
\\
\bottomrule

\end{tabular}
}
\caption{Ablations. We evaluate the success rate on two environments when removing important components of our world model and planner. For the world model modifications, we also report metrics for predictive accuracy explained in the text. All results are averaged over $3$ seeds.}
\label{tab:ablation}
\end{table*}

In particular, given a planning horizon $K$, we first collect a random trajectory $(s_t, a_t)_{t=1 \dots L}$ of length $L=20$  and extract latent embeddings $\{z_t\}_{t=1, \dots L}$ through the encoder $h_\theta$.
We then autoregressively estimate the embedding $z_{K+1}$ using only the first embedding $z_1$ and the action sequence $(a_t)_{t=1 \dots K}$, obtaining a prediction $\hat{z}_{K+1}$.

We repeat this process for $N$ trajectories, obtaining $N$ sequences of latent embeddings $(z^n_t)_{t=1 \dots L}^{n=1 \dots N}$ and $N$ predictions $\{\hat{z}^n_{K+1}\}^{n=1 \dots N}$.
We compute $\texttt{rank}(\hat{z}^n_{K+1})$ as the lowest $k$ such that $\hat{z}^n_{K+1}$ is in the $k$-nearest neighbors of $z^n_{K+1}$ considering all other embeddings $\{z^n_t\}_{t=1, \dots L}$.
We can finally compute $H@K = \frac{1}{N}\sum_{n=1}^{N}\mathbf{1}_{\texttt{rank}(\hat{z}^n_{K+1})=1}$ and $\text{MMR@K} = \frac{1}{N}\sum_{n=1}^{N}\frac{1}{\texttt{rank}(\hat{z}^n_{K+1})}$.

We found that training an inverse model is crucial for learning good representations.
Even if the uninformative loss $L_{\text{margin}}$ introduced in Equation \ref{eq:loss-margin} already helps with avoiding the collapse in the latent space, we were not successful in training the forward model to high predictive accuracy unless the inverse model was jointly trained, despite further hyperparameter tuning.
We can hypothesize that the inverse model enforces a regular structure in the latent space, which is in turn helpful for training the rest of the world model.

On the other hand, we find that the contribution of an hybrid forward model is more environment-dependent.
We ablate this by introducing a forward model that only takes the state embedding $z_t$ and an action $a_t$ as input to predict the next embedding $z_{t+1}$, without having access to a contextual observation $s_c$.
In this case, the forward model can be implemented as a layer-normalized MLP with $3$ layers of $256$ neurons. When adopting this fully latent forward model, predictive accuracy drops sensibly.
While success rate also drops sharply in DigitJump, this is not the case for ProcgenMaze.
We believe that this can be motivated by the fact that several levels of ProcgenMaze could be solved in few steps, only knowing the local structure of the maze with respect to the starting position of the agent.
In DigitJump, on the other hand, the agent can easily move across the environment and needs global information to plan accurately.

\paragraph{Planner}

Evaluating the planning algorithm does not require retraining the world model. We first show the importance of the lookup table.
Without correcting the world model's inaccuracies, the planner is not able to recover from incorrect trajectories.
As a result, the success rate is comparable to that of one-shot planning.
Finally, we empirically validate the importance of state reidentification: when disabled, the BFS procedure is forced to randomly discard new vertices due to the exponential increase in the size of the graph.
Because of this, promising trajectories cannot be expanded and the planner is only effective on simple levels which only require short-term planning.

\begin{figure}[!htb]
\minipage{0.3\textwidth}
  \includegraphics[width=\linewidth]{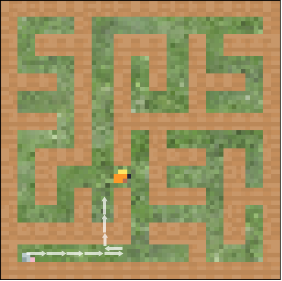}
\endminipage\hfill
\minipage{0.3\textwidth}
  \includegraphics[width=\linewidth]{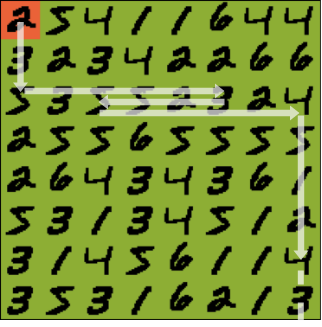}
\endminipage\hfill
\minipage{0.3\textwidth}%
  \includegraphics[width=\linewidth]{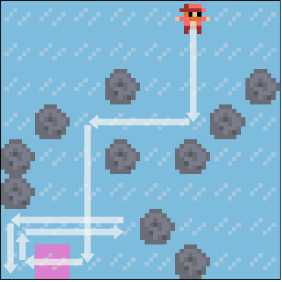}
\endminipage \\
\minipage{0.3\textwidth}
  \includegraphics[width=\linewidth]{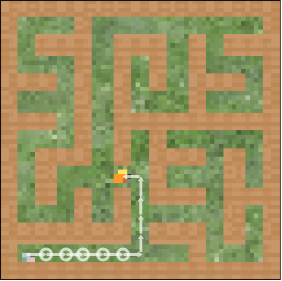}
\endminipage\hfill
\minipage{0.3\textwidth}
  \includegraphics[width=\linewidth]{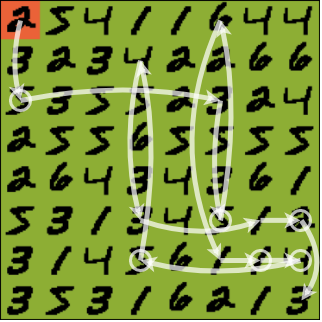}
\endminipage\hfill
\minipage{0.3\textwidth}%
  \includegraphics[width=\linewidth]{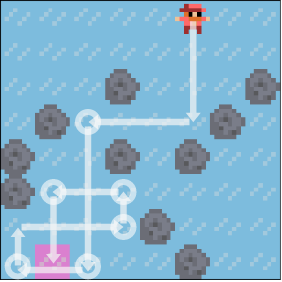}
\endminipage
\caption{Failure cases. On the first row, a level from each environment that one-shot PPGS fails to solve (the white arrows represent the policy). On the second row, the policies corrected by a full planner, which is able to solve all levels. A white circle is drawn when \textsc{PPGS} recomputes its policy.}
\label{fig:failure}
\end{figure}

\paragraph{Failure Cases}
We now present a visual rendition of failure cases for one-shot \textsc{PPGS} in \fig{fig:failure}, together with the correct policy retrieved by the full planner.

\paragraph{Iterative Model Improvement}
In general settings, collecting training trajectories by sampling actions uniformly at random does not grant sufficient coverage of the state space.
In this case, the planners designed for PPGS can be deployed for data collection while training the world model.
To show the potential of this approach in our current setting, we train our method starting from a small number of random trajectories and iteratively collecting on-policy data.
We compare the performance of one-shot planning on IceSlider when trained on 1k levels. We measure success rates on unseen levels for (A) the default setting (400k random training samples), (B) in a low-data scenario (100k random training samples), and (C) when iteratively adding on-policy transitions to a small initial training set of 100k random training samples.
In this last case, we collect 100k additional on-policy samples every 5 epochs.
At 20 epochs we observe that on-policy data collection is able to accelerate and stabilize learning (see \tab{tab:on-policy}).
When training is completed (40 epochs), (A) and (C) reach the same performance, while (B) does not improve.
We believe that this hints that
\begin{itemize}
    \item random trajectories are enough to eventually cover the state space of IceSlider well and,
    \item the planner can be effectively used for data collection and iterative improvements.
\end{itemize}
We therefore believe that collecting data on-policy in an RL-like loop is crucial in environments requiring more exploration and represents an interesting direction.

\begin{table*}[h]
\small
\centering
\begin{tabular}{@{}lccc@{}}
\toprule
& \multicolumn{3}{c}{\textbf{IceSlider} - One-shot success \% on unseen levels}\\
\cmidrule{2-4} & 
(A) 400k offline & (B) 100k offline & (C) 100k off + 300k on-policy \\
\midrule
20 epochs & 0.331$\pm$0.27 & 0.040$\pm$0.01 & \textbf{0.778$\pm$0.07}
\\
40 epochs & 0.826$\pm$0.05 & 0.043$\pm$0.01 & \textbf{0.857$\pm$0.04}
\\

\bottomrule
\end{tabular}
\caption{Performance when training with datasets of various sizes and with on-policy data collection. Collecting trajectories generated by the planner can accelerate learning.}
\label{tab:on-policy}
\end{table*}

\section{Choice of Baselines and Fairness}
\label{app:fairness}

After introducing the fundamental reasons behind our choice of baselines in \sec{sec:experiments}, we present our reasoning and experimental setup with respect to each of the methods.

\paragraph{\textsc{PPO}} \textsc{PPO} \cite{shulman2017ppo} is a strong model-free RL algorithm. Unlike \textsc{PPGS}, \textsc{PPO} requires a reward signal instead of visual goals. We grant \textsc{PPO} an advantageous setting by allowing on-policy data collection for 50M environment steps, which is in stark contrast to the offline dataset of 400k random transitions that \textsc{PPGS} is trained on.
We use the implementation and hyperparameters presented by \citet{cobbe2020procgen} for the Procgen suite, due to its similarity to the rest of the environments. While \textsc{PPGS} is tuned on ProcgenMaze and keeps its hyperparameters fixed across environments, we favor \textsc{PPO} by tuning the number of timesteps per rollout according to the environment to account for the possibility of getting stuck in a funnel state.

\paragraph{\textsc{GLAMOR}} \textsc{GLAMOR} \cite{paster2020planning} learns inverse dynamics to achieve visual goals in Atari games. Similarly to \textsc{PPGS}, \textsc{GLAMOR} does not require a reward signal but needs to receive a visual goal. The only difference with \textsc{PPGS} in terms of settings is that we allow \textsc{GLAMOR} to collect data on-policy and for more interactions (2M). At evaluation time we deploy a strictly more forgiving scheme for \textsc{GLAMOR}, which is described in the original paper \cite{paster2020planning}. As \textsc{GLAMOR} is designed to approximately reach its goals, we also accept trajectories that terminate near the actual goal as viable solutions. Hyperparameters for \textsc{GLAMOR} were tuned by the original authors in Atari, which is a visually comparable setting.

\paragraph{DreamerV2} DreamerV2 \cite{hafner2020mastering} is a model-based RL approach reaching state-of-the-art performance in discrete games and continuous control domains. We use the original implementation for Atari environments. Due to its large computational requirements, we are only able to run DreamerV2 for a reduced number of steps, totaling 4M. We remark that while this is not enough for performance on Atari to converge, it is shown by the original authors to be sufficient for solving a significant number of games.

All cited codebases that we use are publicly available under a MIT license.

\section{Further Implementation Details}
\label{app:further_implementation}

In this section we report relevant implementation choices for \textsc{PPGS}. In any case, we refer to our code \cite{website} for precise details.

\subsection{World Model}\label{sec:implementation}

\paragraph{Encoder.}
The encoding function $h_\theta$ is learned by a convolutional neural network.
The output of the convolutional backbone of a ResNet-18 \cite{he2015deep} is fed through a single fully connected layer with $d$ units, where $d=16$ is the size of the latent space $Z$. The output of the network is normalized to unit L2 norm.

\paragraph{Forward Model.}

From an architectural perspective, our hybrid forward model transforms the state embedding $z_t$ through a deconvolutional network and concatenates it to the RGB observation $s_c$ and a batchwise one-hot tensor representing the action.
The result is processed through a second ResNet-18 to predict the next embedding.
We found it to be irrelevant whether to train the network to predict a state representation $z_{t+1}$ or a latent offset $z_{t+1} - z_{t}$: for our experiments we choose the former.

Similarly to \citet{hafner2019dream} we find that, in practice, explicitly encouraging representations to be predictive for longer horizons (for instance through a multi-step loss) does not appear to be helpful. For this reason, we only train for one-step predictions, as noted in Equation \ref{eq:loss-fw}.

\paragraph{Inverse Model.}
To enforce a simpler structure in the latent space, we implement the inverse model  $p_\omega$ as a low-capacity one-layer MLP with ReLU activations, $32$ neurons and layer normalization \cite{ba2016layer}.

\paragraph{Hyperparameters}
\label{app:hyperparams}

Hyperparameters are fixed in all experiments unless explicitly mentioned. More importantly, we deploy the same set of hyperparameters across all three environments, after tuning them on ProcgenMaze via grid search.
The latent margin $\varepsilon$ is set to $0.1$ and the dimensionality of the latent space $d$ is set to 16.
The world model we propose is optimized using Adam \cite{kingma2014adam} with learning rate $\lambda = 0.001$ for all components and parameters $\varepsilon=0.00001, \beta_1=0.9, \beta_2=0.999$.
All components are trained for 40 epochs with a batch size of 128. The losses are combined as in Equation \ref{eq:loss-total} with weights $\alpha=10, \beta=1$, although our architecture shows robustness to this choice.
Training the world model takes approximately 20 hours on a single NVIDIA ampere GPU.

\subsection{Planner}
\label{app:planner}

In this subsection, we present both planners (one-shot and full) more in detail.

\paragraph{One-shot Planner}
Algorithm \ref{alg:open-loop} offers a more thorough description of the one-shot planner introduced in Algorithm \ref{alg:simplified}. Given a visual goal $g$ and an initial observed state $s$, maximizing discounted rewards corresponds to recovering the shortest action sequence $(a_i)_{1, ...,n}$ such that $s_n = g$.

For this purpose, Algorithm \ref{alg:open-loop} efficiently builds and searches the latent graph.
It has access to an initial high-dimensional state $s$ and a visual goal $g$; it keeps track of a list of visited vertices $V$ and of a set of leaf vertices $L$.
For each visited latent embedding, the algorithm stores the action sequence leading to it in a dictionary $D$.
A key part of the algorithm is represented by the \texttt{filter} function. 
The \texttt{filter} function receives as input the new set of leaves $L'$, from which vertices reidentifiable with visited states have already been discarded.
The function removes elements of $L'$ until no pair of states is too close.
This is done by building a secondary graph with the elements of $L'$ as vertices and edges between all pairs of vertices at less than $\frac{\epsilon}{2}$ distance.
A set of non-conflicting elements can then be recovered by approximately solving a minimum vertex cover problem.
If the state space of the environment grows exponentially with the planning horizon, or if the world model fails to reidentify bisimilar states, $L'$ can still reach impractically large sizes. For this reason, after resolving conflicts, if its cardinality is larger than a cutoff $C=256$, $|L_{t+1}| - C$ elements are uniformly sampled and removed.

\begin{algorithm}[htbp]
\caption{One-shot \textsc{PPGS}}
\textbf{Input:} $s, g$ \\
\textbf{Output:} action sequence $(a_i)_{1, ..., n}$
\begin{algorithmic}[1]
\State $z, z_g = h_\theta(s_1), h_\theta(g)$ \Comment{project to latent space}
\State $V = L = \{z\}$
\State $D = \{z: []\}$
\While{for $T_{\text{MAX}}$ steps}
    \State $L' = \emptyset$
    \For{$z \in L$} \Comment{grow the latent graph}
        \For{$a \in A$}
            \State $z' = f_\phi(z, a, s)$
            \If{$\min_{v \in V} \|z'-v\|_2 > \frac{\varepsilon}{2}$} \Comment{skip if already visited}
                \State{$L' = L' \cup \{z'\}$}
                \State{$D[z'] = D[z] + [a]$}
            \EndIf
        \EndFor
    \EndFor
    \State{$L = \texttt{filter}(L')$} \Comment{select the largest group of elements such that no pair is too close}
    \State{$z ^ \star = \argmin_{z \in L}\|z - z_g\|_2$}
    \If{$\|z^\star - z_g\|_2 \leq \frac{\varepsilon}{2}$} \Comment{if $z^\star$ can be reidentified with the goal}
        \State{return $D[z^\star]$}
    \EndIf
    \State $V = V \cup L$ \Comment{add leaves to visited set}
\EndWhile
\end{algorithmic}
\label{alg:open-loop}
\end{algorithm}

\paragraph{Full Planner}

The full planner used by \textsc{PPGS} introduces the possibility of online replanning in an MPC approach. It autoregressively computes a latent trajectory $T$ conditioned on the action sequence $P$ retrieved by one-shot planning. At each step, the current observation is projected to the latent space to check if it can be reidentified with the predicted embedding in $T$. When this is not possible, the action sequence $P$ is recomputed. Moreover, the planner gradually fills a latent transition buffer $B$. Forward predictions are then computed according to $\hat{f_\theta}(z, a, s)$, which returns $z'$ if $(z, z', a) \in B$, otherwise it queries the learned forward model $f_\theta$. 
As a side note, when replanning while using the full planner, the planning horizon $T_{\text{max}}$ is set to $10$ steps.
We report the method in full in Algorithm \ref{alg:closed-loop}.

\begin{algorithm}[h]
\caption{\textsc{PPGS} (full planner)}
\textbf{Input:} $s, g$
\begin{algorithmic}[1]
\State $z = h_\theta(s)$
\State $B = \emptyset$ \Comment{set of observed latent transitions}
\State $P = \texttt{one\_shot\_\textsc{PPGS}}(s, g)$ \Comment{retrieve initial policy}
\State $T = [z]$
\For{a in P}:
    \State{$T = T + [\hat{f_\phi}(T[-1], a, s)]$}  \Comment{autoregressively predict latent trajectory T}
\EndFor
\State $T.\text{pop}(0)$ \Comment{discard first embedding}
\While{$s \neq g$}
    \State $a = P.\text{pop}(0)$  \Comment{take first action and remove it from the action list}
    \State take action $a$ and reach state $s'$
    \State $z' = h_\theta(s')$
    \State $B = B \cup \{(z, z', a)\}$
    \State $z_\text{pred} = T.\text{pop}(0)$ \Comment{retrieve predicted embedding}
    \If{$\| z_{\text{pred}} -  z'\|_2 > \frac{\varepsilon}{2}$ or A = []} \Comment{if the latent trajectories does not match predictions}
        \State $A = \texttt{one\_shot\_PPGS}(s', g)$ \Comment{replan}
        \State $T = [z]$
        \For{a in P}:
            \State{$T = T + [\hat{f_\phi}(T[-1], a, s)]$}  \Comment{autoregressively predict latent trajectory T}
        \EndFor
        \State $T.\text{pop}(0)$ \Comment{discard first embedding}
    \EndIf
    \State $s = s'$
    \State $z = z'$
\EndWhile
\end{algorithmic}
\label{alg:closed-loop}
\end{algorithm}

\subsection{\textsc{GS on Images}}
\label{app:classical}
Our baselines include a graph search algorithm in observation space that does not involve any learned component. We refer to this algorithm as \textit{\textsc{GS on Images}} and it can be seen as a measure of how hard an environment is when relying on reconstructing the state diagram to solve it. \textsc{GS on Images} assumes solely on the deterministic nature of the environment. Given a starting state $s$, a goal state $g$ and the action set $A$, \textsc{GS on Images} plans as shown in Algorithm \ref{alg:classical}. It relies on a dictionary $A_{\text{left}}$ which stores, for each visited state, the set of actions that have not been attempted yet, and on a graph representation of the environment $G=(V, E)$, where $V$ is the set of visited states and $E$ contains the observed transitions between states and labeled by an action.

As a side note on this algorithm's performance, we remark that, unlike the remaining methods, it strongly depends on the absence of visual noise and distractors.
In particular, this method is bound to fail in environments in which this assumption does not hold: visual noise would render reidentification meaningless for GS on Images and the baseline would not be able to avoid revisiting vertices of the latent graph. 

\begin{algorithm}[htbp]
\caption{\textsc{GS on Images}}
\textbf{Input:} $s, g$
\begin{algorithmic}[1]
\State $A_{\text{left}} = \{s: A\}$
\State $V = \{s\}$
\State $E = \emptyset$
\While{$s \neq g$}
    \If {$A_{\text{left}}[s] = \emptyset$}
        
        \If {$\exists s' \in V$ such that $A_{\text{left}}[s'] \neq \emptyset$ and $s'$ is reachable from $s$}
            \State {find and apply action sequence to reach closest $s' \in V$ }
            \State $s = s'$
        \Else
            \State \textbf{return}
        \EndIf
    \Else
        \State $a \sim \mathcal{U}(A_{\text{left}}[s])$ \Comment{uniformly sample among remaining actions}
        \State $A_{\text{left}}[s] = A_{\text{left}}[s] \setminus \{a\}$
        \State take action $a$ and reach state $s'$
        \If {$s' \notin V$}
            \State $V = V \cup \{s'\}$
            \State $A_{\text{left}}[s] = A$
        \EndIf
        \State $E = E \cup (s, s', a)$
    \EndIf
\EndWhile
\end{algorithmic}
\label{alg:classical}
\end{algorithm}

\subsection{Data Collection}

As our method solely relies on accurately modeling the dynamics of the environment, the only requirement for training data is sufficient coverage of the state space.
In most cases, this is satisfied by collecting trajectories offline according to a uniformly random policy.
The ability to leverage a fixed dataset of trajectories draws \textsc{PPGS} closer to off-policy methods or even batch reinforcement learning approaches.

In practice, unless specified otherwise, we collect $20$ trajectories of $20$ steps from a set of $n=1000$ training levels, for a total of $400k$ environment steps.
One exception is made for ProcgenMaze, for which we also set a random starting position at each episode, since uncorrelated exploration is not sufficient to cover significant parts of the state.

\section{Environments}
\label{app:environments}

In this section, we present a few remarks on the environments chosen. For ProcgenMaze, we choose what is reported as the \textit{easy} distribution in \citet{cobbe2020procgen}. This corresponds to grids of size $n \times n$, with $3 \leq n \leq 15$; each cell of the grid is either a piece of wall or a corridor. 
In IceSlider, the agent always starts in the top row and needs to descend to a goal on the bottom row. It is not sufficient to slide over the goal, but the agent needs to come to a full stop on the correct square. In DigitJump, the handwritten digits are the same across training and test levels. Their frequency and position does of course change.

All environments return observations as 64x64 RGB images. ProcgenMaze and IceSlider are rendered in a similar style to ATARI games, while the DigitJump is a grid of MNIST \cite{lecun1998gradient} digits that highlights the cell at which the agent is positioned.
The action space in all cases is restricted to four cardinal actions (UP, DOWN, LEFT, RIGHT) and a no-op action, for a total of $5$ actions.

\begin{figure}[htbp]
  \centering
\resizebox{\textwidth}{!}{
\begin{tabular}{rl@{\hspace{0.05cm}}l@{\hspace{0.05cm}}l@{\hspace{0.05cm}}l@{\hspace{0.05cm}}l}
ProcgenMaze & 
\includegraphics[width=.1\linewidth]{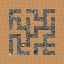} &
\includegraphics[width=.1\linewidth]{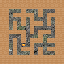} & \includegraphics[width=.1\linewidth]{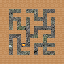} & \includegraphics[width=.1\linewidth]{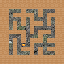} & \includegraphics[width=.1\linewidth]{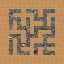} \\
&
\includegraphics[width=.1\linewidth]{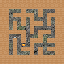} & \includegraphics[width=.1\linewidth]{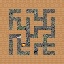} & \includegraphics[width=.1\linewidth]{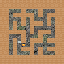} & \includegraphics[width=.1\linewidth]{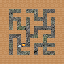} & \includegraphics[width=.1\linewidth]{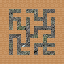} \\
\\
DigitJump & 
\includegraphics[width=.1\linewidth]{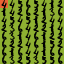} &
\includegraphics[width=.1\linewidth]{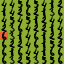} & \includegraphics[width=.1\linewidth]{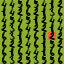} & \includegraphics[width=.1\linewidth]{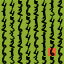} & \includegraphics[width=.1\linewidth]{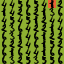}\\
&
\includegraphics[width=.1\linewidth]{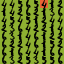} & \includegraphics[width=.1\linewidth]{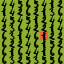} & \includegraphics[width=.1\linewidth]{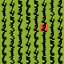} & \includegraphics[width=.1\linewidth]{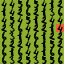} & \includegraphics[width=.1\linewidth]{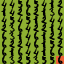} \\
\\
IceSlider & 
\includegraphics[width=.1\linewidth]{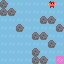} &
\includegraphics[width=.1\linewidth]{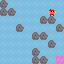} & \includegraphics[width=.1\linewidth]{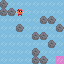} & \includegraphics[width=.1\linewidth]{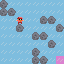} & \includegraphics[width=.1\linewidth]{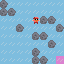}\\
&
\includegraphics[width=.1\linewidth]{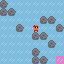} & \includegraphics[width=.1\linewidth]{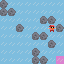} & \includegraphics[width=.1\linewidth]{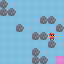} & \includegraphics[width=.1\linewidth]{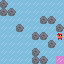} & \includegraphics[width=.1\linewidth]{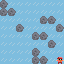} \\

\end{tabular}
}
\caption{Expert trajectories for a level extracted from each of the environments.}
\label{fig:trajectories}
\end{figure}

Examples of expert trajectories are shown in \fig{fig:trajectories}. For more information on the environments, we refer the reader to our code \cite{environments}.

\section{Numerical Results}
We finally include the full numerical results from \fig{fig:success} and \fig{fig:generalization} in \tab{tab:numerical}.

\begin{table*}[h]
\small
\centering
\resizebox{\columnwidth}{!}{%
\begin{tabular}{@{}lccccccccccccc@{}}
\toprule
& \multicolumn{13}{c}{\textbf{ProcgenMaze} - Success \% on unseen levels when training on $n$ levels}\\
\cmidrule{2-14} & 
$n$=1 & 2 & 5 & 10 & 20 & 50 & 100 & 200 & 500 & 1000 & 2000 & 5000 & 10000\\
\midrule
PPGS  & 0.320 & 0.447 & 0.530 & 0.607 & 0.703 & 0.743 & 0.770 & 0.783 & 0.843 & 0.897 & 0.850 & 0.880 & 0.880
\\
PPO  & 0.115 & 0.113 & 0.139 & 0.168 & 0.203 & 0.321 & 0.417 & 0.455 & 0.577 & 0.667 & 0.746 & 0.853 & 0.880
\\
DreamerV2 & - & - & - & - & - & - & - & - & - & 0.153 & - & - & -
\\
GLAMOR & - & - & - & - & - & - & - & - & - & 0.100 & - & - & -
\\

\bottomrule
\\
& \multicolumn{13}{c}{\textbf{IceSlider} - Success \% on unseen levels when training on $n$ levels}\\
\cmidrule{2-14} & 
$n$=1 & 2 & 5 & 10 & 20 & 50 & 100 & 200 & 500 & 1000 & 2000 & 5000 & 10000\\
\midrule
PPGS  & 0.223 & 0.273 & 0.296 & 0.260 & 0.347 & 0.450 & 0.550 & 0.573 & 0.873 & 0.970 & 0.960 & 0.965 & 0.965
\\
PPO  & 0.004 & 0.000 & 0.032 & 0.115 & 0.179 & 0.260 & 0.274 & 0.298 & 0.421 & 0.564 & 0.565 & 0.590 & 0.601
\\
DreamerV2 & - & - & - & - & - & - & - & - & - & 0.007 & - & - & -
\\
GLAMOR & - & - & - & - & - & - & - & - & - & 0.010 & - & - & -
\\

\bottomrule
\\
& \multicolumn{13}{c}{\textbf{DigitJump} - Success \% on unseen levels when training on $n$ levels}\\
\cmidrule{2-14} & 
$n$=1 & 2 & 5 & 10 & 20 & 50 & 100 & 200 & 500 & 1000 & 2000 & 5000 & 10000\\
\midrule
PPGS  & 0.160 & 0.170 & 0.157 & 0.157 & 0.220 & 0.247 & 0.193 & 0.250 & 0.790 & 0.990 & 0.960 & 0.940 & 0.980
\\
PPO  & 0.107 & 0.137 & 0.109 & 0.161 & 0.291 & 0.331 & 0.405 & 0.502 & 0.441 & 0.475 & 0.513 & 0.488 & 0.492
\\
DreamerV2 & - & - & - & - & - & - & - & - & - & 0.193 & - & - & -
\\
GLAMOR & - & - & - & - & - & - & - & - & - & 0.133 & - & - & -
\\

\bottomrule
\end{tabular}
}
\caption{Generalization results. This table presents the numerical results used to produce \fig{fig:success} and \fig{fig:generalization}. All metrics are averaged over 3 random seeds.}
\label{tab:numerical}
\end{table*}

\end{document}